# Equity forecast: Predicting long term stock price movement using machine learning

Nikola Milosevic
School of Computer Science,
University of Manchester, UK
Nikola.milosevic@manchester.ac.uk

## Abstract

Long term investment is one of the major investment strategies. However, calculating intrinsic value of some company and evaluating shares for long term investment is not easy, since analyst have to care about a large number of financial indicators and evaluate them in a right manner. So far, little help in predicting the direction of the company value over the longer period of time has been provided from the machines. In this paper we present a machine learning aided approach to evaluate the equity's future price over the long time. Our method is able to correctly predict whether some company's value will be 10% higher or not over the period of one year in 76.5% of cases.

## 1 INTRODUCTION

Trading stocks on the stock market is one of the major investment activities. In the past, investors developed a number of stock analysis method that could help them predict the direction of stock price movement. Modelling and predicting of equity future price, based on the current financial information and news, is of enormous use to the investors. Investors want to know whether some stock will rise or fall over certain period of time. In order to predict how some company, in which investor want to invest, would perform in future, they developed a number of analysis methods based on current and past financial data and other information about the company. Financial balance sheets and various ratios that describe the health of company are the bases of technical analysis that investors undertake to analyze and predict company's future stock prize. Predicting the direction of stock price is particularly important for value investing.

Experienced analysts could apply some mathematical models that are proven based on the past data in order to evaluate company's intrinsic value, such as Graham number. Graham number and Graham's criteria is probably one of the most famous models (Graham, 1949). However, due to the increased volatility in the current market, it would be probably impossible to find a company that satisfies Graham's principles on today's stock exchanges. Because of these changes, the need for adjusted models rose. Also, stock market changes over time (Barsky & De Long, 1992; Hendershott & Moulton, 2011). New investment strategies and new technology were introduced, which made some of the old models obsolete. Since financial literacy became higher, there are more market players than ever. However, for some of the old models cannot be easily adopted for the changes in stock market.

The introduction to algorithms in trading definitely changed the stock market. Algorithms made it easy to react fast to certain events on the stock market. Machine learning algorithms also enabled analysts to create models for predicting prices of stocks much easier. Introduction of machine learning caused that new models can be developed based on the past data. In this paper we will describe the method for predicting stock market prices using several machine learning algorithms. Our main hypothesis was that by applying machine learning and training it on the past data, it is possible to predict the movement of the stock price, as well as the ratio of the movement over certain fixed amount of time. Particularly, we want to determine stocks that will rise over 10% in a period of one year. The percentage of growth or fall in a stock price can be variable, however, in order to make our case we will focus on growth of 10%. So far, investors usually were not able to predict the percentage of stock price growth over the time. However, this information is vital for investors while selecting stocks. Usually, investors want to select the stocks which will grow in price substantially.

The paper is organized in the following manner: In Section 2, we will review relevant works for this topic from the past. In Section 3, we will describe our method and algorithms we used. Section 4 describes the results of our method on our dataset. In the last section we will discuss the method, its results, compare it to other methods and propose some future work.

## 2 RELEVANT WORKS

Wilson and Sharda studied prediction firm bankruptcy using neural networks and classical multiple discriminant analysis, where neural networks performed significantly better than multiple discriminant analysis (Wilson & Sharda, 1994). Min and Lee were doing prediction of bankruptcy using machine learning. They evaluated methods based on SVM, multiple discriminant analysis, logistic regression analysis, and three-layer fully connected back-propagation neural networks. They results indicated that support vector machines outperformed outer approaches (Min & Lee, 2005). Similarly, Tam was predicting bank bankruptcy using neural networks (Tam, 1991).

Lee was trying to predict credit rating of a company using support vector machines. They used various financial indicator and ratios such as interest coverage ratio, ordinary income to total assets, Net income to stakeholders' equity, current liabilities ratio, etc. and achieved accuracy of around 60% (Lee, 2007). Predicting credit rating of the companies were also studied using neural networks achieving accuracy between 75% and 80% for US and Taiwan markets (Huang, et al., 2004).

Phua et al. performed a study predicting movement of major five stock indexes: DAX, DJIA, FTSE-100, HSI and NASDAQ. They used neural networks and they were able to predict the sign of price movement with accuracy higher than 60% by using component stocks as input for the prediction (Phua, et al., 2003).

SVM was applied to predict the direction of time series. Kim (Kim, 2003) trained SVM on daily time series from of Korean stock market. He reported a hit rate of around 56%. Huang et al did a tried to use support vector machines in order to predict weekly movement of Japanese NIKKEI 225 index. Their approach achieved 73% hit rate with SVM and 75% with combined model. Also SVM outperformed in their approach backpropagation neural networks (Huang, et al., 2005). Schumaker and Chen examined the prediction that could be achieved by analysis of textual articles using SVM. They were trying to predict the movement of S&P 500 index 20 minutes after the release of the news. They used several approaches such as bag of words, noun phrases and named entities and achieved directional hit rate of 57% (Schumaker & Chen, 2009). Analysis of the textual news, however, is not in scope of this research.

Tsai and Wang did a research where they tried to predict stock prices by using ensemble learning, composed of decision trees and artificial neural networks. They created dataset from Taiwanese stock market data, taking into account fundamental indexes, technical indexes, and macroeconomic indexes. The performance of DT+ANN trained on Taiwan stock exchange data showed F-score performance of 77%. Single algorithms showed F-score performance up to 67% (Tsai & Wang, 2009).

To the best of his knowledge, the author is not aware of any work on predicting equity prices on the stock market after longer time period. We are interested in long term investment, where the investor holds the stock for extended period of time. In the following sections we will examine the machine learning aided method for analyzing and predicting growth of certain stock over the extended time period.

# 3 METHOD

## 3.1 DATASET

The dataset was obtained by using Bloomberg terminal. We selected stocks from indexes such as S&P 1000, FTSE 100 and S&P Europe 350. In total we selected 1739 stocks. For each stock we obtained the stock price at the end of each quarter from the first quarter of 2012 until the final quarter of 2015. Along the price, we have also retrieved the following financial indicators about each company in our dataset:

**Book value** - the net asset value of a company, calculated by total assets minus intangible assets (patents, goodwill) and liabilities.

**Market capitalization** - the market value of a company's issued share capital; it is equal to the share price times the number of shares outstanding.

**Change of stock Net price over the one month period**

**Percentage change of Net price over the one month period**

**Dividend yield** - indicates how much a company pays out in dividends each year relative to its share price.

**Earnings per share** - a portion of a company's profit divided by the number of issued shares. Earnings per share serves as an indicator of a company's profitability.

**Earnings per share growth –** the growth of earnings per share over the trailing one-year period.

**Sales revenue turnover -**

**Net revenue -** the proceeds from the sale of an asset, minus commissions, taxes, or other expenses related to the sale.

**Net revenue growth** – the growth of Net revenue over the trailing one-year period.

**Sales growth** – sales growth over the trailing one-year period.

**Price to earnings ratio** – measures company's current share price relative to its per-share earnings.

**Price to earnings ratio, five years average** – averaged price to earnings ratio over the period of five years.

**Price to book ratio -** compares a company's current market price to its book value.

**Price to sales ratio –** ratio calculated by dividing the company's market cap by the revenue in the most recent year.

**Dividend per share -** is the total dividends paid out over an entire year divided by the number of ordinary shares issued.

**Current ratio** - compares a firm's current assets to its current liabilities.

**Quick ratio** - compares the total amount of cash, marketable securities and accounts receivable to the amount of current liabilities.

**Total debt to equity** - ratio used to measure a company's financial leverage, calculated by dividing a company's total liabilities by its stockholders' equity.

**Analyst ratio –** ratio given by human analyst.

**Revenue growth adjusted by 5 year compound annual growth ratio**

**Profit margin –** a profitability ratio calculated as net income divided by revenue, or net profits divided by sales

**Operating margin -** ratio used to measure a company's pricing strategy and operating efficiency. It is a measurement of what proportion of a company's revenue is left over after paying for variable costs of production such as wages, raw materials, etc.

**Asset turnover** - the ratio of the value of a company's sales or revenues generated relative to the value of its assets[1].

## 3.2 PREDICTING EQUITY PRICE MOVEMENT METHODOLOGY

We modelled out task of predicting equity price movement as classification task, in which we classify stocks that will have 10% higher price in one year period as "Good" ones and others as "Bad". Since we have historical data retrieved from Bloomberg terminal, we created a dataset which had indicator values and price on some history date more than one year ago. We created a script in Python that was comparing history price with the price exactly one year after the first price was measured. If the price was 10% higher, the script would label data point as "Good", otherwise as "Bad".

For creating a good machine learning model is required balanced dataset (Ganganwar, 2012; Akbani, et al., 2004). In order to create such a dataset we selected only the same number of stocks labeled as "Good" as stocks labeled as "Bad". Since our dataset contained 1739 stocks and we needed to discard some stocks because they would imbalance our dataset, our dataset contained 1298 data rows (649 labelled as Good and 649 labelled as Bad). These data rows contained indicators of the stocks from the final quarter of 2014. and label was made based on comparison of prices in final quarter of 2015. and 2014. However, since we had historical data, we created additional data rows looking at the quarter before the final one in 2014., and its price one year later. Our final data set contained 4538 data rows.

---
[1] Definition of ratios and indicators are from http://www.investopedia.com/ and http://lexicon.ft.com/

Since not all financial indicators were available for all companies in our data set, we assigned value -9999 to not present or not available values.

We used several classification machine learning algorithms to train the model in Weka toolkit (Hall, et al., 2009). We trained the models using C4.5 decision trees, Support Vector Machines with Sequential Minimal Optimization, JRip, Random Trees, Random Forest, Logistic regression, Naïve Bayes and Bayesian Networks. Firstly we performed 10-fold cross validation on all these algorithms with all indicators and history price used as features. Then we performed manual feature selection by removing features and evaluating whether performance of the algorithm improved or decreased. We performed this process iteratively, until we didn't get the optimal model with minimal number of features and the best performance.

# 4 RESULTS

The results achieved with all features (financial indicators) during 10-fold cross-validation are presented in Table 1.

*Table 1: Results of machine learning prediction based on all financial indicators*

| Algorithm | Precision | Recall | F-score |
|---|---|---|---|
| C4.5 decision trees | 0.687 | 0.687 | 0.687 |
| SVM with SMO | 0.639 | 0.630 | 0.624 |
| JRip | 0.635 | 0.635 | 0.635 |
| Random Tree | 0.668 | 0.668 | 0.668 |
| **Random Forest** | **0.751** | **0.751** | **0.751** |
| Logistic regression | 0.643 | 0.637 | 0.633 |
| Naïve Bayes | 0.545 | 0.530 | 0.487 |
| Bayesian Networks | 0.625 | 0.618 | 0.612 |

Algorithm that performed the best was Random Forests with Precision, Recall and F-score of 75.1%. We performed also a paired t-test with 0.05 significance level. Random Forests performed significantly better than other algorithms.

In order to improve the performance of random forest tried to select the features in the optimal way. We performed feature selection manually, selecting certain set of features and reevaluating the results. At the end of this iterative process, 11 features were selected that showed the best performance. Original feature set had 28 features. While some of the features did not affect or affect negatively the final results, the facts some ratios were indicating were duplicated in other ratios, so they were unnecessary.

Performance for the Random Forests with only 11 features that contributed to performance in our tests were 0.762 for precision, recall and F-Score. The percentage of the correctly classified instances was 76.2%, in the experiment evaluated using 10-fold cross validation. The features that we selected are presented in Table 2.

Table 2: Selected features that showed the best performance

| Attribute |
| --- |
| book_value |
| market_cap |
| DIVIDEND_YIELD |
| BEST_EPS |
| PE_RATIO |
| PX_TO_BOOK_RATIO |
| BEST_DPS |
| CUR_RATIO |
| QUICK_RATIO |
| TOT_DEBT_TO_TOT_EQY |
| history_price |

We evaluated also other algorithms, we previously used on the dataset with selected features. The results are presented in Table 3.

Table 3: The results of machine learning based equity prediction using 10-fold cross validation and 11 selected features

| Algorithm | Precision | Recall | F-score |
| --- | --- | --- | --- |
| C4.5 decision trees | 0.660 | 0.660 | 0.660 |
| SVM with SMO | 0.636 | 0.629 | 0.624 |
| JRip | 0.640 | 0.639 | 0.639 |
| Random Tree | 0.700 | 0.700 | 0.700 |
| **Random Forest** | **0.765** | **0.765** | **0.765** |
| Logistic regression | 0.638 | 0.630 | 0.625 |
| Naïve Bayes | 0.526 | 0.515 | 0.453 |
| Bayesian Networks | 0.641 | 0.626 | 0.615 |

With the selected features, Random Forests remained the best performing algorithm and they improved slightly by 1.4%. This is not huge improvement, but it proves that the proper selection of feature can improve the results of equity prediction.

## 5 DISCUSSION

In this paper is presented a machine learning aided methodology for equity movement prediction over the long time. With all 28 selected financial indicators, the methodology performs with F-score of 75.1%, however, by doing a feature selection the number of features can be reduced to 11, while performance can be increased to F-score of 76.5%. Although the increase is not large, the algorithm is more efficient and faster with smaller set of features.

Some of the features from the larger set were not necessary, since they were not giving any relevant information about company's valuation, while the others were just duplicating the fact told by already analyzed financial indicator. For example, it is possible to assume the value of earnings if the value of total stock number and earning per share ratio is available.

It seems that information about growth is not necessary. From this is could be deduced that ratios and information that describes current financial state of the company, without a look at the past performances is enough for predicting future behavior of the company (with accuracy showed in this work). This principle can be especially useful for investors that want to invest in new companies. Hypothesis that companies can be valued and their future can be predicted only by looking at present data has to be further tested, however, it proved to be correct in our case for our dataset. We will leave this hypotheses to be tested by future researchers in more details.

Our algorithm performed with 76.5% F-score, correctly classifying 76.5% instances. This means that the tool that provide this model to the investor can significantly help investor in his decision making by recommending him/her the stocks that are probable to perform well. Algorithm still makes mistakes and labels some stocks that will perform well as the ones that will perform badly and vice-versa. However, to the best of our knowledge, this is state of the art performance for the long term equity price direction prediction.

The limitation of this study is that the models are not created out of data that were not limited in time. More accurate way of generating financial machine learning models would be to limit training data until certain year, while test it on unseen future data. In this case, that has not been done and whole dataset is split into training and testing folds and tested against. Markets may change over time, and therefore n-fold cross-validation evaluation would not capture these changes and may introduce, what is in finance called "Look-Ahead Bias". Look-ahead bias occurs when the study may assume that information was available to investors at a point in time, when in fact the information was not yet publicly available (Davis, 1994).